\documentclass[conference]{IEEEtran}

\usepackage{amsmath,amssymb,amsfonts,mathrsfs,bm}
\usepackage{amsthm}
\usepackage{cite}
\usepackage{array}

\usepackage[shortlabels]{enumitem}
\usepackage{graphicx}
\usepackage{url}
\usepackage{color}
\usepackage{algorithm,algpseudocode}

\usepackage{multirow}
\usepackage[table]{xcolor}
\usepackage[normalem]{ulem}
\usepackage{array}
\newcolumntype{L}[1]{>{\raggedright\let\newline\\\arraybackslash\hspace{0pt}}m{#1}}
\newcolumntype{C}[1]{>{\centering\let\newline\\\arraybackslash\hspace{0pt}}m{#1}}
\newcolumntype{R}[1]{>{\raggedleft\let\newline\\\arraybackslash\hspace{0pt}}m{#1}}
\usepackage{subfigure}
\usepackage{xparse}
\usepackage[capitalize]{cleveref}

\newcommand{\Real}{\mathbb{R}}
\newcommand{\ba}{\mathbf{a}}

\newcommand{\bb}{\mathbf{b}}

\newcommand{\bF}{\mathbf{F}}

\newcommand{\bG}{\mathbf{G}}

\newcommand{\bH}{\mathbf{H}}

\newcommand{\bI}{\mathbf{I}}

\newcommand{\bK}{\mathbf{K}}

\newcommand{\bM}{\mathbf{M}}

\newcommand{\bp}{\mathbf{p}}
\newcommand{\bP}{\mathbf{P}}

\newcommand{\bQ}{\mathbf{Q}}
\newcommand{\br}{\mathbf{r}}
\newcommand{\bR}{\mathbf{R}}

\newcommand{\bS}{\mathbf{S}}

\newcommand{\bT}{\mathbf{T}}

\newcommand{\bv}{\mathbf{v}}

\newcommand{\bw}{\mathbf{w}}

\newcommand{\bchi}{\bm{\chi}}
\newcommand{\bxi}{\bm{\xi}}
\newcommand{\bepsilon}{\bm{\epsilon}}
\DeclareMathOperator*{\argmin}{arg\,min}
\crefname{Remark}{Remark}{Remarks}
\newtheorem{Remark}{Remark}

\begin{document}

\title{UWB/LiDAR Fusion For Cooperative Range-Only SLAM }

\author{\IEEEauthorblockN{Yang Song\IEEEauthorrefmark{1},
Mingyang Guan\IEEEauthorrefmark{1}, 
Wee Peng Tay,
Choi Look Law, and 
Changyun Wen}
\IEEEauthorblockA{School of Electrical and Electronic Engineering, Nanyang Technological University\\
\IEEEauthorrefmark{1} Two authors contributed equally to this work.}}
\maketitle



\maketitle \thispagestyle{empty}


\begin{abstract}
We equip an ultra-wideband (UWB) node and a 2D LiDAR sensor a.k.a. 2D laser rangefinder on a mobile robot, and place UWB beacon nodes at unknown locations in an unknown environment. All UWB nodes can do ranging with each other thus forming a cooperative sensor networks. We propose to fuse the peer-to-peer ranges measured between UWB nodes and laser scanning information, i.e. range measured between robot and nearby objects/obstacles, for simultaneous localization of the robot, all UWB beacons and LiDAR mapping. The fusion is inspired by two facts: 
1) LiDAR may improve UWB-only localization accuracy as it gives more precise and comprehensive picture of the surrounding environment; 
2) on the other hand, UWB ranging measurements may remove the error accumulated in the LiDAR-based SLAM algorithm. 
Our experiments demonstrate that UWB/LiDAR fusion enables drift-free SLAM in real-time based on ranging measurements only.

\end{abstract}


\section{Introduction}
\label{sect:intro}

Simultaneous localization and mapping, also known as SLAM, has attracted immense attention in the mobile robotics literature, and many approaches use laser range finders (LiDARs) due to their ability to accurately measure range to the nearby objects. There are two basic approaches to mapping with LiDARs: feature extraction and scan matching. The scan matching approach matches point clouds directly and relate two robot's poses via map constraints.  Scan matching is much more adaptable than feature extraction approach as it is much less environmentally dependent. Of all existing scan matching algorithms, GMapping \cite{grisettiyz2005improving} and HectorSLAM \cite{kohlbrecher2011flexible} are arguably the most well-known and widely used algorithms.  GMapping needs odometry input while HectorSLAM is an odometry-less approach. Apart from their respective advantages, one common drawback is that their performance is very vulnerable to the accumulated errors, which may come from long run operation of odometry such as GMapping which extracts odometry information directly from odometry sensor or the SLAM algorithm itself such as HectorSLAM where the error of robot's pose at current time step will be passed via scan matching procedure to the grid-map which will in turn impair the estimation of robot's pose at next time step.

To eliminate these accumulated errors, loop-closure detection approach \cite{HessICRA2016} is introduced to detect whether the robot has returned to a previously visited location. This approach relies heavily on accuracy of the robot's pose estimate.  One another way to remove accumulated errors and enhance the robustness in LiDAR-based SLAM is by sensor fusion \cite{wang2017ultra}. In this paper, we propose to fuse LiDAR sensor with UWB sensors. 
The objective of fusing UWB and LiDAR is to 1) provide not just landmarks/beacons but also a detailed mapping information about surrounding environment; 2) improve the accuracy and robustness of UWB-based localization and mapping.

\begin{figure}[!htb]
\centering
\includegraphics[width=1\linewidth]{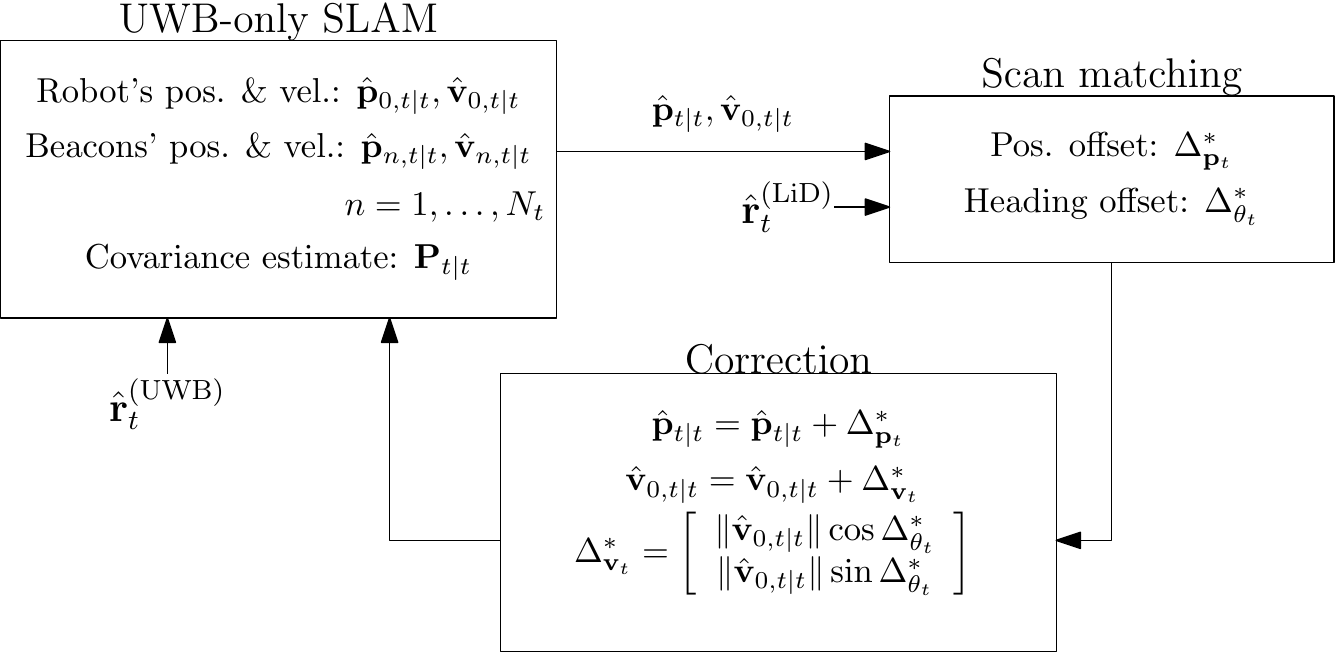}
\caption{The digram of proposed system. }
\label{fig:system}
\end{figure}

However, the fusion is hindered by the discrepancy between the accuracy of the UWB mapping and that of LiDAR mapping. 
UWB has lower range resolution than LiDAR (the laser ranging error is about 1cm which is about one-tenth of UWB ranging error) so that UWB cannot represent an environment in the same quality as LiDAR can do. In this case, fusion by building LiDAR map directly on top of UWB localization results is not a good solution. In our proposed fusion scheme, we consider to refine the robot's pose obtained from UWB ranging measurements by subjecting it to a scan matching procedure. 
The diagram of our proposed system is shown in \cref{fig:system} where the system collects all peer-to-peer UWB ranging measurements consisting of robot-to-beacon ranges and beacon-to-beacon ranges at time $t$, based on which the robot and beacons' 2D positions and 2D velocities are estimated using extended Kalman filter (EKF), then system feeds the state estimate to scan matching procedure in order to update map and find state offset, which is used to correct EKF's state estimate.  

Why do we choose UWB? In many practical environments such as enclosed areas, urban canyons, high-accuracy localization becomes a very challenging problem  in the presence of multipaths and NLOS propagations. Among various wireless technologies such as the Bluetooth, the Wi-Fi, the UWB and the ZigBee, UWB is the most promising technology to combat multipaths in cluttered environment. The ultra-wide bandwidth in UWB results in well separated direct path from the multi-paths, thus enabling more accurate ranging using time-of-arrival (TOA) of the direct path.

\subsection{Related Work}
\label{sect:related}

\subsubsection{Range-only SLAM}
\cite{DjugashICRA2006,BlancoICRA2008,DeiIPIN2010,VallicrosaIROS2014,HerranzICRA2014,GeneveIROS2015,TiemannICCW2018,GentnerPLANS2018} propose to simultaneously localize robot(s) and static beacons based on only robot-to-beacon ranging measurements aided by control input e.g. odometry/IMU/etc..  
Among these works, \cite{DeiIPIN2010,TiemannICCW2018,GentnerPLANS2018} use UWB technology to measure ranges.
\cite{DjugashICRA2006,DjugashER2008, BlancoICRA2008, MenegattiICRA2009} consider robot-to-beacon as well as beacon-to-beacon ranges. Our work further extends the paradigm of \cite{BlancoICRA2008} by integrating the robot-to-obstacle ranges obtained by laser range finder. The new paradigm allows us to not just localize the robot(s) and map the landmarks (i.e. localize the beacons) but also map the obstacles around the robot. 
Moreover, as the robot's pose can be estimated based on UWB ranging measurements, the heading information can be derived from its estimated trajectory thus we need no control input.

\subsubsection{LiDAR based SLAM}
Well-known approaches such as HectorSLAM\cite{kohlbrecher2011flexible}, GMapping\cite{grisettiyz2005improving} have pushed forward the research on LiDAR-based SLAM. Gmapping proposes an adaptive approach to learn grid maps using RaoBlackwellized particle filter (RBPF) while the number of required particles in RBPF can be dramatically reduced. HectorSLAM proposes a fast online learning of occupancy grid maps using a fast approximation of map gradients and a multi-resolution grid. Moreover, GMapping needs odometry input whereas HectorSLAM purely replies on the laser ranger finder.

\subsubsection{Sensor fusion for SLAM}
As LiDAR cannot see through occulusions such as smoke, fog or dust, \cite{fritsche2017modeling} proposes to fuse LiDAR data with Radar data to handle the SLAM problem in the environments with low visibility. The fusion takes place on scan level and map level to maximize the map quality considering the visibility situation. 
\cite{maddern2016real} proposes a probabilistic model to fuse sparse 3D LiDAR data with stereo images to provide accurate dense depth maps and uncertainty estimates in real-time. 
\cite{wang2017ultra} proposes to fuse UWB and visual inertial odometry to remove the visual drift with the aid of UWB ranging measurements thus improving the robustness of system. This system is anchor-based where our system is anchor-less.
\cite{perez2017multi} proposes a batch-mode solution to fuse ranging measurements from UWB sensors and 3D point-clouds from RGB-D sensors for mapping and localizing unmanned aerial vehicles (UAVs). This system makes the same assumption as ours where the locations of UWB anchors are priori unknown. However, our system can simultaneously localize the robot, UWB beacons and build map in real-time whereas \cite{perez2017multi} proposes to collect all the data before start doing mapping and localization.

\subsection{Our Contributions}
We propose a 2D range-only SLAM approach that combines low-cost UWB sensors and LiDAR sensor for simultaneously localizing the robot, beacons and constructing a 2D map of an unknown environment in real-time. Here are some highlights of the proposed approach: 
\begin{enumerate}
\item  no prior knowledge about the robot's initial position is required, neither is control input;
\item  beacon(s) may be moved and number of beacon(s) may be varied while SLAM is proceeding;  
\item  the robot can move fast without accumulating errors in constructing map;
\item  the map can be built even in feature-less environment.
\end{enumerate}

{\em Notations}: We use $\Real$ to denote the set of real numbers, the superscript $^T$ to represent matrix transpose, and the superscript $^{-1}$ to denote matrix inverse. We use $\bI_n$ to represent an identity matrix of size $n\times n$ and use ${\bf 1}_n$ to denote a column vector with $n$ elements all being one. We use $\otimes$ to denote kronecker product, use $|\cdot|$ to take absolute value, and use $\|\cdot\|$ to denote Euclidean distance. The normal distribution with mean $\mu$ and variance $\sigma^2$ is denoted as ${\cal N}(0,\sigma^2)$. We use $\left[{\bf A}\right]_i$ to denote the $i$th row of matrix ${\bf A}$, 
and use $\left[{\bf A}\right]_{i,j}$ to denote the element at the intersection of the $i$th row and the $j$th column of matrix ${\bf A}$.

\section{Problem Formulation}
\label{section:pf}

We consider to fuse two kinds of ranging measurements which are 
\begin{itemize}
\item the peer-to-peer ranges between UWB nodes consisting of robot-to-beacon ranges and beacon-to-beacon ranges, and 
\item the range between the robot and obstacles available from laser range finder.
\end{itemize} 
We symbolize UWB-based ranging measurements as ${\bf r}^{(\rm UWB)}$ and LiDAR-based ranging measurements as ${\bf r}^{(\rm LiD)}$. Let 
\begin{align*}
{\bf x}_t =& \left[p_{0,x,t}, p_{0,y,t}, v_{0,x,t}, v_{0,y,t}\right]^T, \\
{\bf m}^{(\rm UWB)}_t =& \left[p_{1,x,t}, p_{1,y,t}, \ldots, p_{N_t,x,t}, p_{N_t,y,t}, \right. \\
&\left.v_{1,x,t}, v_{1,y,t},\ldots,v_{N_t,x,t}, v_{N_t,y,t}\right]^T,
\end{align*} 
where ${\bf x}_t \in \Real^{4\time 1}$ contains 2D location of the robot 
$\bp_{0,t}=\left[p_{0,x,t}, p_{0,y,t}\right]^T$ and 2D velocity 
$\bv_{0,t}=\left[v_{0,x,t}, v_{0,y,t}\right]^T$ at time $t$, 
${\bf m}^{(\rm UWB)}_t \in \Real^{4N_t}$ which represents "UWB map" contains $N_t$ number of UWB beacons' 2D locations 
$\bp_{n,t}=\left[p_{n,x,t}, p_{n,y,t}\right]^T, n=1,\ldots,N_t$ and 2D velocities 
$\bv_{n,t}=\left[v_{n,x,t}, v_{n,y,t}\right]^T, n=1,\ldots,N_t$, and let ${\bf m}_t^{(\rm LiD)}$ be the "LiDAR map" containing the locations of the observed obstacles. 
\begin{Remark}
No control input e.g. odometry/IMU/etc., is assumed. The heading information (or equivalently 2D velocity) for robot/UWB beacons is derived purely from their trajectory.
\end{Remark}
Given all ranging measurements ${\bf r}^{(\rm UWB)}_{1:t}$ and ${\bf r}^{(\rm LiD)}_{1:t}$ up to time $t$ , our goal is to simultaneously estimate ${\bf x}_t$, ${\bf m}^{(\rm UWB)}_t$ and ${\bf m}^{(\rm LiD)}_t$. We propose to decompose the problem into three coupled steps:
\begin{enumerate}
\item find the relative positions of robot and beacons based on ${\bf r}^{(\rm UWB)}_{1:t}$, and derive the robot's heading information from its trajectory,
\item construct/update LiDAR map as well as UWB map using ${\bf r}^{(\rm LiD)}_{1:t}$ and robot's pose and heading estimates and beacons' pose estimates, 
\item correct robot's pose and heading as well as beacons' poses based on the map's feedback.
\end{enumerate}
In what follows, we will elaborate how we implement these three steps. 

\section{UWB-Only SLAM}
\label{section:UWB-SLAM}

\subsection{The dynamical and observational models}
\label{sect:dyn_obs}
Let $\bchi_t=\left[\bp_t^T,\bv_t^T\right]^T\in\Real^{4(N_t+1)}$ represent the compete UWB-related state consisting of locations of robot and beacons 
$\bp_t=\left[\bp_{0,t}^T,\ldots,\bp_{N_t,t}^T\right]^T$ and velocities of robot and beacons 
$\bv_t=\left[\bv_{0,t}^T,\ldots,\bv_{N_t,t}^T\right]^T$. The state $\bchi_t$ is evolved according to the following dynamical model:
\begin{align*}
\bchi_{t}= \bF_t \bchi_{t-1} + \bG_t \bw_{t},
\end{align*}
where the transition matrix $\bF_t$ equals to
$
\left[
\begin{array}{cc}
\bI_{2(N_t+1)} & \delta\bI_{2(N_t+1)} \\
{\bf 0} & \bI_{2(N_t+1)}
\end{array}
\right]
$, 
and $\bG_t$ equals to $\left[
\begin{array}{c}
\delta {\bf I}_{2(N_t+1)} \\
{\bf I}_{2(N_t+1)}
\end{array}
\right]$, the state noise $\bw_{t}$ is zero-mean and has covariance $\bQ_{t} = \sigma^2_w \bG_t\bG_t^T$ \cite{LiSPIE2000} and $\delta$ is the sampling interval.

Let ${\bf r}^{(\rm UWB)}_t\in\Real^{N_t(N_t+1)/2}$ be the UWB-based ranges measured at time $t$ that consists of robot-to-beacon ranges  
$\left[{\bf r}^{(\rm UWB)}_t\right]_i,i=1,\ldots,N_t$, and beacon-to-beacon ranges 
$\left[{\bf r}^{(\rm UWB)}_t\right]_i,i=N_t+1,\ldots, N_t(N_t+1)/2$, and they are are non-linear function of $\bp_t$: 
\begin{align*}
\left[{\bf r}^{(\rm UWB)}_t\right]_i =& h\left(\bp_{j,t},\bp_{k,t}\right)=\left\| \bp_{j,t} - \bp_{k,t} \right\|,0\leq j<k\leq N_t,
\end{align*}
where $i=1,\ldots,N_t(N_t+1)/2$ index the pairwise combination of UWB nodes. 
We assume that all peer-to-peer ranges in $\left[{\bf r}_t\right]_i,i=1,\ldots,N_t(N_t+1)/2$ are corrupted by i.i.d. additive noise $n_t\sim {\cal N}(0,\sigma_n^2)$, i.e.,
$\left[\hat{\bf r}^{(\rm UWB)}_t\right]_i = \left[{\bf r}^{(\rm UWB)}_t\right]_i + n_t$. 
This assumption has been widely adopted under line-of-sight (LOS) scenario \cite{WymeerschPIEEE0209}.

\subsection{EKF update}
The update process proceeds via a standard EKF:
\begin{align*}
\hat{\bchi}_{t|t-1} =& \bF_t \hat{\bchi}_{t-1|t-1} + \bw_t, \\
\bP_{t|t-1} =& \bF_t \bP_{t-1|t-1} \bF^T_t + \bQ_t, \\
\bS_t =& \bH_t \bP_{t|t-1} \bH^T_t + \sigma^2_n \bI_{N_t(N_t+1)/2}, \\
\bK_t =& \bP_{t|t-1} \bH^T_t \bS^{-1}_t, \\
\hat{\bchi}_{t|t} =& \hat{\bchi}_{t|t-1} + \bK_t \left(\hat{\br}^{(\rm UWB)}_t - h\left(\hat{\bchi}_{t|t-1}\right)\right), \\
\bP_{t|t} =& \left(\bI_{4(N_t+1)}-\bK_t\bH_t\right)\bP_{t|t-1},
\end{align*}
where $\hat{\bchi}_{t|t}$ and $\bP_{t|t}$ are the updated state estimate and covariance estimate, respectively. The measurement matrix $\bH_t\in \Real^{L_t\times 4(N_t+1)}$ is defined to be the $\left. \frac{\partial h}{\partial \bchi} \right|_{\hat{\bchi}_{t|t-1}}$. As the measurements depend not on the velocity of the nodes, the partial derivatives of $h(\cdot)$ w.r.t. the velocities of robot and beacons are all zeros.
\begin{Remark}
The estimated velocity of the robot won't be accurate if robot stops because it is derived from robot's trajectory. 
\end{Remark}

\subsection{Elimination of location ambiguities}
\label{sect:ambi}
\begin{figure}[!htb]
\centering
\includegraphics[width=.38\linewidth]{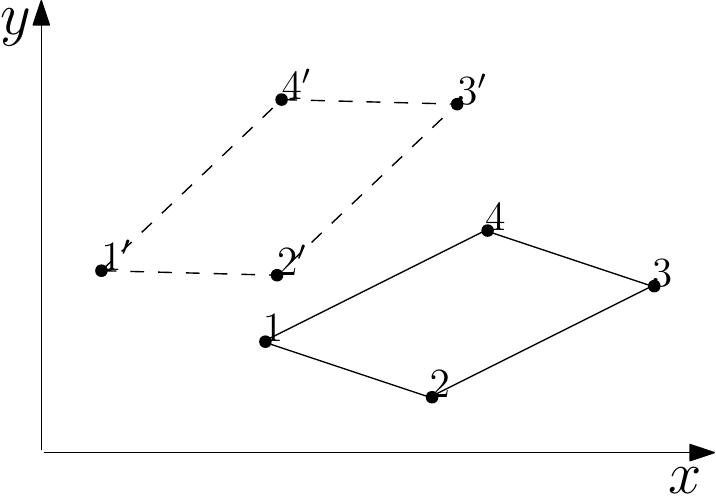}
\caption{Illustration of two versions of the same relative geometry of four nodes: one is translated and rotated version of the other one. }
\label{fig:rela}
\end{figure}

As we are dealing with an infrastructure-less localization problem where no prior information about location of nodes is assumed, we can only derive the relative geometry of all UWB nodes based on peer-to-peer ranging measurements. Such relative geometry, as shown in \cref{fig:rela}, however, can be arbitrarily translated and rotated. 
For constructing "LiDAR map" which will be discussed in next section, we need to eliminate all ambiguous copies of the relative geometry except one. To do this, we translate and rotate the state estimate available $\hat{\bchi}_{t|t}$:
$
\bR_t(\theta)\otimes \bI_2 
\left[\begin{array}{c}
\hat{\bp}_{t|t} \\
\hat{\bv}_{t|t}
\end{array}\right]+
\left[\begin{array}{c}
\bT_t \\
{\bf 0}\end{array}\right]
$, 
where $\bR_t(\theta)$ is a 2D rotation matrix and $\bT_t\in \Real^{2\times (N_t+1)}$ is a translation matrix. We find $\bR_t(\theta)$ and $\bT_t$
such that one beacon node is always located at origin $[0,0]^T$ and one another beacon node is always located on positive $x$-axis. In addition, we force the velocity of these two chosen beacon nodes to be zero. These two beacon nodes will set up a local frame under which the LiDAR map can be built and updated.

\subsection{Detection/removal of ranging outliers}
NLOS propagation is crucial for high-resolution localization systems \cite{DardariJPROC0309,MaranoJSAC0810} because non-negligibly positive biases will be introduced in range measurements, thus degrading the localization accuracy.  A rule-of-thumb suggested by \cite{GururajIPIN2017,dwm1000} has been practised for NLOS detection. This rule says if the power difference, i.e. the received power minus the first-path power, is greater than 10dB the channel is likely to be NLOS. The NLOS ranges are ignored in the EKF update.


\begin{Remark}
We have tried to detect the outliers using hypothesis testing \cite{HolICUWB2009} on the residuals of the EKF, i.e. $\bepsilon_t=\hat{\br}_t^{(\rm UWB)} - h\left(\hat{\bp}_{t|t-1}\right)$, which is a very common way to remove outliers in the infrastructure-based localization system. In absence of errors, $\left[\bepsilon_t\right]_i \sim {\cal N}\left(0, \left[\bS_t\right]_{i,i} \right)$. Given confidence intervals for each peer-to-peer ranging measurement, if it's violated, this range is considered an outlier and is ignored. This detection method doesn't work well in relative positioning because in the relative positioning where localization of all nodes purely relies on peer-to-peer ranges if some ranges measured at time $t-1$ are outliers, the ranging error will be propagated to all nodes whereby the prediction ranges $\hat{\br}_{t|t-1}^{(\rm UWB)}$ may deviate away from the ranges $\hat{\br}_t^{(\rm UWB)}$ measured at time $t$ even if $\hat{\br}_t^{(\rm UWB)}$ contains no outliers thus resulting in many false alarm detections.
\end{Remark}

\section{Map Update}
\label{section:lidar}

Due to the non-negligible accuracy gap in ranging measurement between UWB and LiDAR, constructing LiDAR map directly on top of UWB localization results is not a promising solution. For example, if the robot keeps scanning the same object for a while, all the scan endpoints should converge to the area where the object is located. However, if the robot's pose and heading are subject to some uncertainties, say 20cm and 3 degrees, respectively, which are quite normal in UWB-only SLAM, mapping the scan endpoints according to robot's pose/heading estimates without any further process would end up with scan endpoints being diverged thereby dramatically reducing the map quality. \cref{fig:nsnc} show such an example. This motivates us to align the scan endpoints by refining the robot's pose/heading estimates. Then, we turn our focus to the scan matching procedure. HectorSLAM \cite{kohlbrecher2011flexible} proposes a scan matching procedure based on Gauss-Newton method \cite{lucas1981iterative} that matches the beam endpoints observed at time $t$ with the latest map ${\bf m}_{t-1}^{(\rm LiD)}$ in order to find the optimal transformation/rotation leading to the best match.
The scan matching proceeds in a multi-resolution grid starting from grid maps with lower resolution to the grid maps with higher resolution so that it's more likely to find the global solution other than being trapped in local solutions.  Our proposed scan matching method as will be discussed below is grounded on \cite{kohlbrecher2011flexible}.

The pose estimates of robot and beacons $\hat{\bp}_{t|t}$ and the velocity estimate of the robot $\hat{\bv}_{0,t|t}$, which are obtained from EKF-based UWB-only SLAM, would be used for initializing the scan matching procedure. Note that the robot's heading information $\theta_t$ could be derived from the 2D velocity as  
$\hat{\theta}_t = \arctan\left(\left[\hat{\bv}_{0,t|t}\right]_2, \left[\hat{\bv}_{0,t|t}\right]_1\right)$.

Let $\bxi_t = \left[\left(\hat{\bp}_{t|t}\right)^T, \hat{\theta}_t\right]^T$. We propose to find the optimal offset $\Delta_{\bxi_t}=\left[\Delta_{\bp_t},\Delta_{\theta_t}\right]^T$  such that the following objective is minimized:
\begin{align}
\label{eq:opt}
	&\Delta_{\bxi_t}^* \nonumber \\
	=& 
	\argmin_{\Delta_{\bxi_t}}\frac{1}{2}\sum_{i=1}^{n}\left[1 - \Gamma(f_i\left(\hat{\bp}_{0,t|t}+\Delta_{\bp_{0,t}}, \hat{\theta}_t+\Delta_{\theta_t}, \hat{\br}^{(\rm LiD)}_t\right)\right]^2  \nonumber\\ 
	&+\frac{\gamma}{2} \sum_{0\leq i<j\leq N_t} \left[ 
	h\left(\hat{\bp}_{i,t|t}+\Delta_{\bp_{i,t}},\hat{\bp}_{j,t|t}+\Delta_{\bp_{j,t}} \right)
	 \right. \nonumber\\
	&\hspace{1.7in}\left.- \left[\hat{\br}^{(\rm UWB)}_t\right]_k \right]^2,
\end{align}
where $h(\ba,\bb)=\|\ba-\bb\|$ and $k$ indexes the range measured between node $i$ and node $j$ and  $\Delta_{\bp_t}=\left[\Delta^T_{\bp_{0,t}},\ldots,\Delta^T_{\bp_{N_t,t}}\right]^T$.
\begin{Remark}
As discussed in \cref{sect:ambi}, two UWB beacons are used to establish the UWB coordinate system i.e. one node is always located at $[0,0]^T$ and the other node is fixed on the positive $x$-axis by letting its coordinate along $y$-axis be zero. Hence, the values corresponding to these three coordinates in $\Delta_{\bxi_t}$ are fixed to zero and we don't update them. 
\end{Remark}

Now let us take a close look at \cref{eq:opt}.

{\em LiDAR map matching:} The first term in \cref{eq:opt} intends to match the laser scan with learnt map by offsetting the robot's pose and heading. The $n$ is the number of laser scan endpoints, and the function $f_i(\cdot)$ maps by transforming/rotating the scan endpoints, say $\bp_i^{(\rm sc)},i=1,\ldots,n$ which are computed based on $ \hat{\br}^{(\rm LiD)}_t$, to their corresponding coordinates $\bp_i^{(\rm UWB)}=f_i\left(\bp_i^{(\rm sc)}\right),i=1,\ldots,n$ under UWB coordinate system, and the $\Gamma(\cdot)$ function returns the occupancy probabilities at coordinates $\bp_i^{(\rm UWB)}=[x_i,y_i]^T,i=1,\ldots,n$. The occupancy probability is the probability of a grid cell where $f_i\left(\bp_i^{(\rm sc)}\right)$ is in being occupied. Follow the way proposed in \cite{kohlbrecher2011flexible}, the occupancy probability $\Gamma\left(\bp_i^{(\rm UWB)}\right)\in[0,1]$ for the $i$th scan endpoint is approximated by bilinear interpolation using its four closest integer neighbour points in the grid map, say $\bp_{i,j}=[x_i,y_j]^T,i,j=\{0,1\}$. This bilinear interpolation can be written as
\begin{align*}
	\Gamma\left(\bp_i^{(\rm UWB)}\right) &\approx \dfrac{y_i - y_0}{y_1-y_0} \left(\dfrac{x_i-x_0}{x_1-x_0}\Gamma(\bp_{1,1})+\dfrac{x_1-x_i}{x_1-x_0}\Gamma(\bp_{0,1})\right) \nonumber\\
	& + \dfrac{y_1- y_i}{y_1-y_0}\left(\dfrac{x_i-x_0}{x_1-x_0}\Gamma(\bp_{1,0})+\dfrac{x_1-x_i}{x_1-x_0}\Gamma(\bp_{0,0})\right).
\end{align*}
The gradient of $\nabla{\Gamma\left(f_i(\bxi_t)\right)}$ is $\left[\dfrac{\partial \Gamma\left(f_i(\bxi_t)\right)}{\partial x_i},\dfrac{\partial \Gamma\left(f_i(\bxi_t)\right)}{\partial y_i}\right]^T$. For the detailed computation of the gradient, we refer the readers to  \cite{kohlbrecher2011flexible}.

\begin{Remark}
The UWB-only SLAM defines a relative coordinate system (\cref{sect:ambi} explains how it's done) on which the LiDAR map is built/updated. That's why the scan endpoints are all mapped to UWB coordinate system.
\end{Remark}

{\em UWB map matching:} The second term in \cref{eq:opt} is a non-linear least square function depending on all LOS peer-to-peer UWB ranging measurements. Minimizing this function would refine the robot and beacons location estimates where $\gamma$ is a tradeoff parameter. When $\gamma=0$, \cref{eq:opt} degenerates to (7) in \cite{kohlbrecher2011flexible} where the matching is done based on LiDAR only.

To find closed-form solution of \cref{eq:opt},  we approximate functions 
$\Gamma(\bxi_t+\Delta_{\bxi_t})$ and $h(\bxi_t+\Delta_{\bxi_t})$, respectively, by first-order Taylor expansion at point ${\bxi_t}$ as
$\Gamma(\bxi_t+\Delta_{\bxi_t})\approx \Gamma(\bxi_t)+\Delta^T_{\bxi_t} \nabla{\Gamma(\bxi_t)}$ and 
$h(\bxi_t+\Delta_{\bxi_t})\approx h(\bxi_t)+\Delta^T_{\bxi_t} \nabla{h(\bxi_t)}$. Then, taking derivative of the objective in \cref{eq:opt} w.r.t. $\Delta_{\bxi_t}$ and equating it to zero yields
\begin{align}
\label{eq:delta_opt}
	\Delta_{\bxi_t}^* = \left(\bH^{(\rm LiD)}_t - \gamma\bH^{(\rm UWB)}_t\right)^{-1} \left(\bM^{(\rm LiD)}_t + \gamma\bM^{(\rm UWB)}_t\right),
\end{align}
where 
\begin{align*}
	\bH^{(\rm LiD)}_t =& \sum_{i=1}^{n}\left[\nabla{\Gamma\left(f_i(\bxi_t)\right)}\dfrac{\partial f_i(\bxi_t)}{\partial \bxi_t}\right]
	\left[\nabla{\Gamma\left(f_i(\bxi_t)\right)}\dfrac{\partial f_i(\bxi_t)}{\partial \bxi_t}\right]^T,\\
	\bH^{(\rm UWB)}_t =& \sum_{0\leq i < j\leq N_t}\nabla{h\left(\hat{\bp}_{i,t|t}+\Delta_{\bp_{i,t}},\hat{\bp}_{j,t|t}+\Delta_{\bp_{j,t}}\right)} \\&\hspace{.5in}\nabla{h\left(\hat{\bp}_{i,t|t}+\Delta_{\bp_{i,t}},\hat{\bp}_{j,t|t}+\Delta_{\bp_{j,t}}\right)}^T,\\
	\bM^{(\rm LiD)}_t =& \sum_{i=1}^{n}\left[\nabla{\Gamma\left(f_i(\bxi_t)\right)}\dfrac{\partial f_i(\bxi_t)}{\partial \bxi_t}\right]\left[1-\Gamma\left(f_i(\bxi_t)\right)\right],\\
	\bM^{(\rm UWB)}_t =& \sum_{0\leq i < j\leq N_t} \nabla{h\left(\hat{\bp}_{i,t|t}+\Delta_{\bp_{i,t}},\hat{\bp}_{j,t|t}+\Delta_{\bp_{j,t}}\right)}\\ &\left(h\left(\hat{\bp}_{i,t|t}+\Delta_{\bp_{i,t}},\hat{\bp}_{j,t|t}+\Delta_{\bp_{j,t}}\right)-\left[\hat{\br}_t^{(\rm UWB)}\right]_k\right),
\end{align*}
where $k$ indexes the range measured between node $i$ and node $j$.
Note that the optimal offset $\Delta_{\bxi_t}^*$ in \cref{eq:delta_opt} generalized the one given in (12) of \cite{kohlbrecher2011flexible} which considers only the LiDAR map matching.

After obtaining the optimal offset $\Delta_{\bxi_t}^*=\left[\left(\Delta_{\bp_t}^*\right)^T,\Delta_{\theta_t}^*\right]^T$, the state estimate would be corrected as
\begin{align*}
\hat{\bf p}_{t|t}=&\hat{\bf p}_{t|t}+\Delta^*_{{\bf p}_{t}}, \\
\hat{\bf v}_{0,t|t}=&\hat{\bf v}_{0,t|t}+\Delta^*_{{\bf v}_{t}}, \\
\Delta_{{\bf v}_{t}}=&
\left[
\begin{array}{c}
\|\hat{\bf v}_{0,t|t}\|\cos\Delta^*_{\theta_{t}}\\
\|\hat{\bf v}_{0,t|t}\|\sin\Delta^*_{\theta_{t}}
\end{array}
\right].
\end{align*}

%

\section{Experimental Results}
\label{section:experiment}

We present here the experimental results that show the advantages of UWB/LiDAR fusion in SLAM by comparing our approach with the existing LiDAR-only SLAM approach. The robustness of UWB/LiDAR fusion in SLAM are demonstrated under different scenarios such as 1) the robot moves fast; 2) beacon(s) drop from the sensor networks / new beacon(s) join the sensor networks / beacons are moving; 3) the environment is feature-less. We also study the impact of $\gamma$, which tradeoffs the LiDAR map matching and UWB map matching, on the SLAM performance. All the experiments shown below share some common settings: the standard deviation (std.) of ranging/motion noise are $\sigma_n=0.1$ meter and $\sigma_w=0.1$ meter.

\subsection{Hardware}
\begin{figure}[!htb]
\centering
\mbox{
\subfigure[UWB hardware platform. ]{
\includegraphics[width=.48\linewidth]{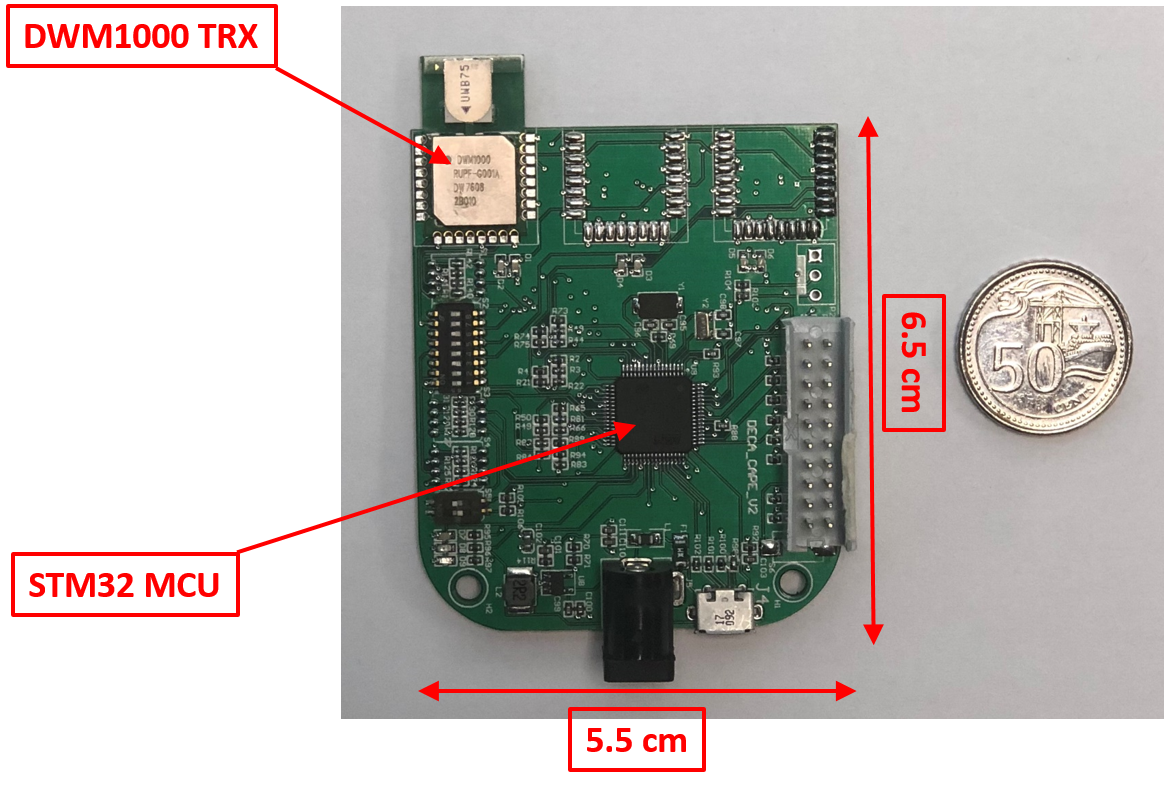}
\label{fig:uwb}}
\subfigure[Robot]{
\includegraphics[width=.43\linewidth]{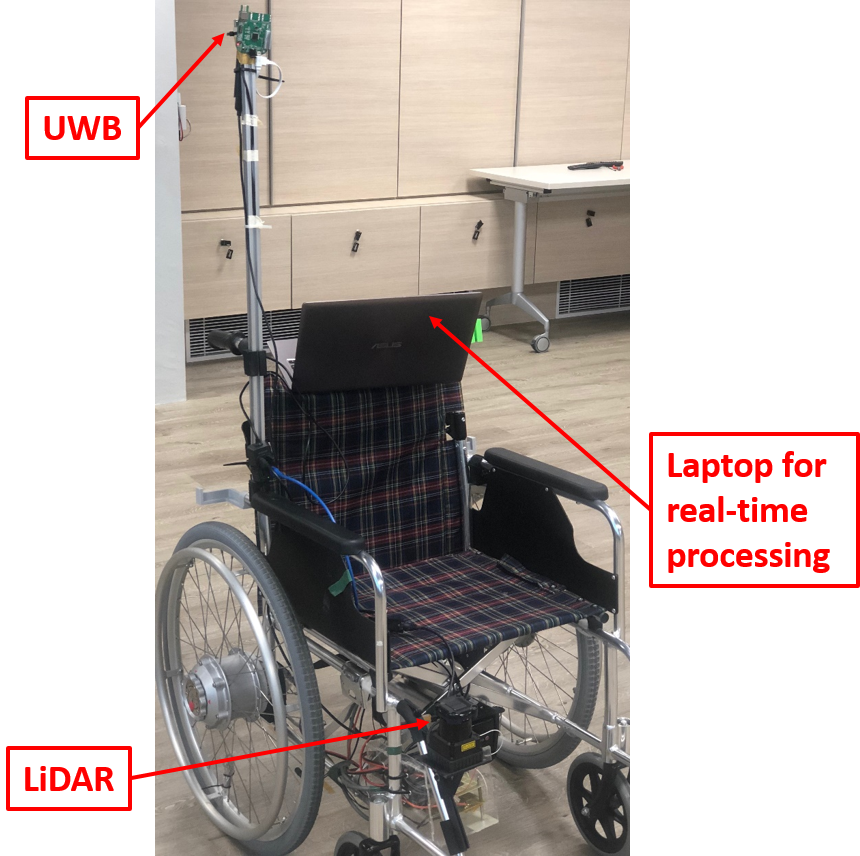}
\label{fig:robot}}}
\caption{Our UWB/LiDAR SLAM system.}
\end{figure}

\cref{fig:uwb} is our designed UWB hardware platform that integrates DWM1000 from Decawave which is used for wireless ranging and messaging. 
The current implementation uses 4GHz band, 850 kbps data rate at 16MHz PRF. The STM32 is controlled by the intelligence of computer/micro-computer e.g.  raspberry pi, beaglebone black, etc., to execute commands given via USB. For one peer-to-peer ranging, it takes 3 milliseconds. The LiDAR sensor we use is Hokuyo UTM-30LX-EW scanning laser rangefinder. The robot i.e. a wheelchair equipped with one UWB sensor and one LiDAR sensor is shown \cref{fig:robot}. The wheelchair is pushed manually in our experiments.

\subsection{SLAM in a workshop of size $12\times19 ~{\rm m}^2$}
\begin{figure}[!htb]
\centering
\mbox{
\subfigure[no scan matching/correction]{
\includegraphics[width=.44\linewidth]{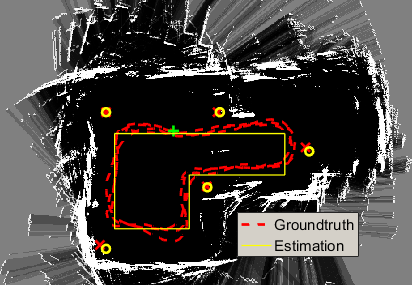}
\label{fig:nsnc}}
\subfigure[$\gamma=0.65$, no correction]{
\includegraphics[width=.44\linewidth]{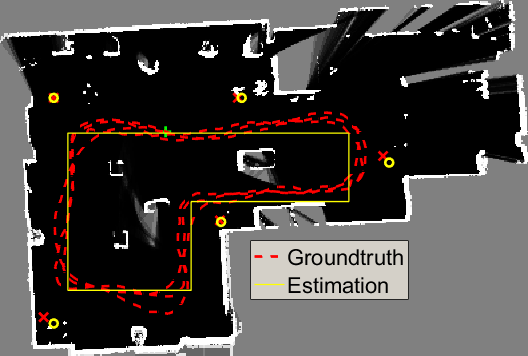}
\label{fig:psnc}}}
\mbox{
\subfigure[$\gamma=0.65$, w/ correction]{
\includegraphics[width=.44\linewidth]{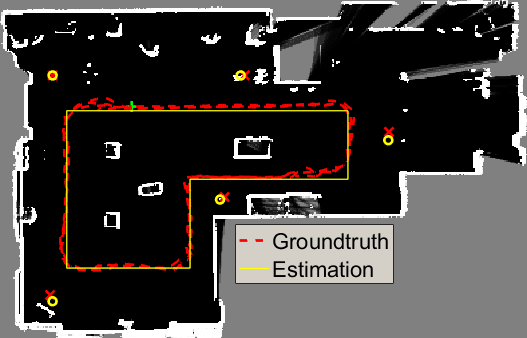}
\label{fig:pspc}}
\subfigure[$\gamma=10^{-6}$, w/ correction]{
\includegraphics[width=.44\linewidth]{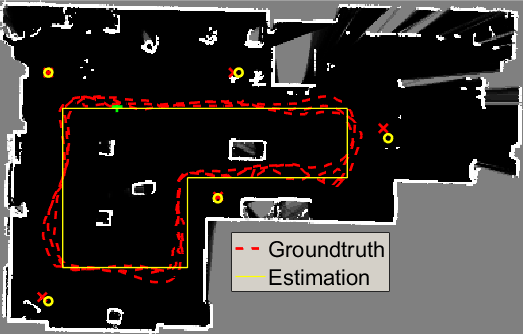}
\label{fig:pspc_lidar}}
}
\caption{UWB/LiDAR fusion in four cases: 
a) no scan matching, no correction; 
b) with UWB/LiDAR scan matching where $\gamma=0.65$, no correction; 
c) with UWB/LiDAR scan matching where $\gamma=0.65$ and correction;
d) with LiDAR-only scan matching where $\gamma=10^{-6}$ and correction. The green "+" denotes the final position of the robot.
}
\label{fig:fusion1}
\end{figure}

Five UWB beacons are placed at unknown locations and the robot moves at a speed about 0.8m/s along a predefined trajectory for three loops which takes about 150 seconds. \cref{fig:fusion1} shows the UWB/LiDAR-based SLAM results. Comparing \cref{fig:nsnc} with \cref{fig:psnc,fig:pspc,fig:pspc_lidar}, we see that when there is no scan matching step the quality of the LiDAR map dramatically decreases. This explains why we cannot build LiDAR map by directly mapping the laser scan endpoints to the UWB map. Comparing \cref{fig:psnc} and \cref{fig:pspc} where the former one has no correction step (i.e. correction of the UWB map using the offset obtained in the scan matching procedure) whereas the latter one has, we can see the correction step significantly improves the estimate of robot's trajectory while the estimates of beacons' positions are roughly the same for both cases. It makes sense since the robot's pose is directly affected by LiDAR map matching, i.e. the first term in \cref{eq:opt}, which however has no direct influence on beacons' poses. \cref{fig:pspc_lidar} ignores the UWB map matching, i.e. the second term in \cref{eq:opt}, by setting $\gamma$ to be a negligibly small value $10^{-6}$. Comparing \cref{fig:pspc_lidar} with \cref{fig:pspc}, we can see the localization for both robot and beacons are distorted when the role of second term in \cref{eq:opt} is downplayed.  
\cref{tab:4cases} gives the averaged (over time steps and over five nodes) positioning error and std. of UWB beacons. The estimates of five beacons' poses are recorded while robot is moving. From the table we can see choosing $\gamma=0.65$ and enabling the correction step gives best results among the three settings. In the following experiments, we empirically choose $\gamma=0.65$ but we believe the choice of $\gamma$ is environmentally dependent. A simple rule-of-thumb is that let $\gamma$ be large when the environments have less features and/or there are sufficient LOS UWB ranges, and let $\gamma$ be small when the environments are feature-sufficient and/or there are only few LOS UWB ranges. 
\begin{table}[!htb]
\vspace{-.1in}
\centering
\caption{Averaged errors/stds. of five UWB beacons' pose estimates.}
\begin{tabular}{c|c|c|c}
& $\gamma=0.65$ & $\gamma=0.65$ & $\gamma=10^{-6}$ \\
& no correction & w/ correction & w/ correction \\
       \hline
  err. in meters & 0.213 & 0.076 & 0.206 \\
  std. in meters & 0.136 & 0.122 & 0.149   
\end{tabular}
\label{tab:4cases}
\vspace{-.1in}
\end{table}

\begin{figure}[!htb]
\vspace{-.15in}
\centering
\mbox{
\subfigure[$135 \leq t\leq 311$]{
\includegraphics[width=.33\linewidth]{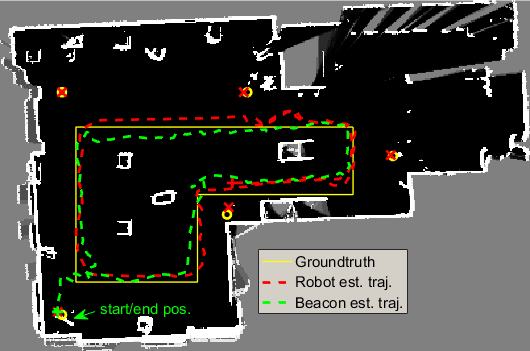}
\label{fig:slot1}}
\hspace{-0.1in}
\subfigure[$466\leq t\leq 626$]{
\includegraphics[width=.33\linewidth]{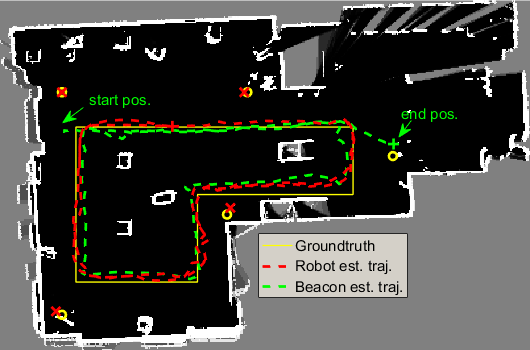}
\label{fig:slot2}}
\hspace{-0.1in}
\subfigure[$N_t$]{
\includegraphics[width=.31\linewidth]{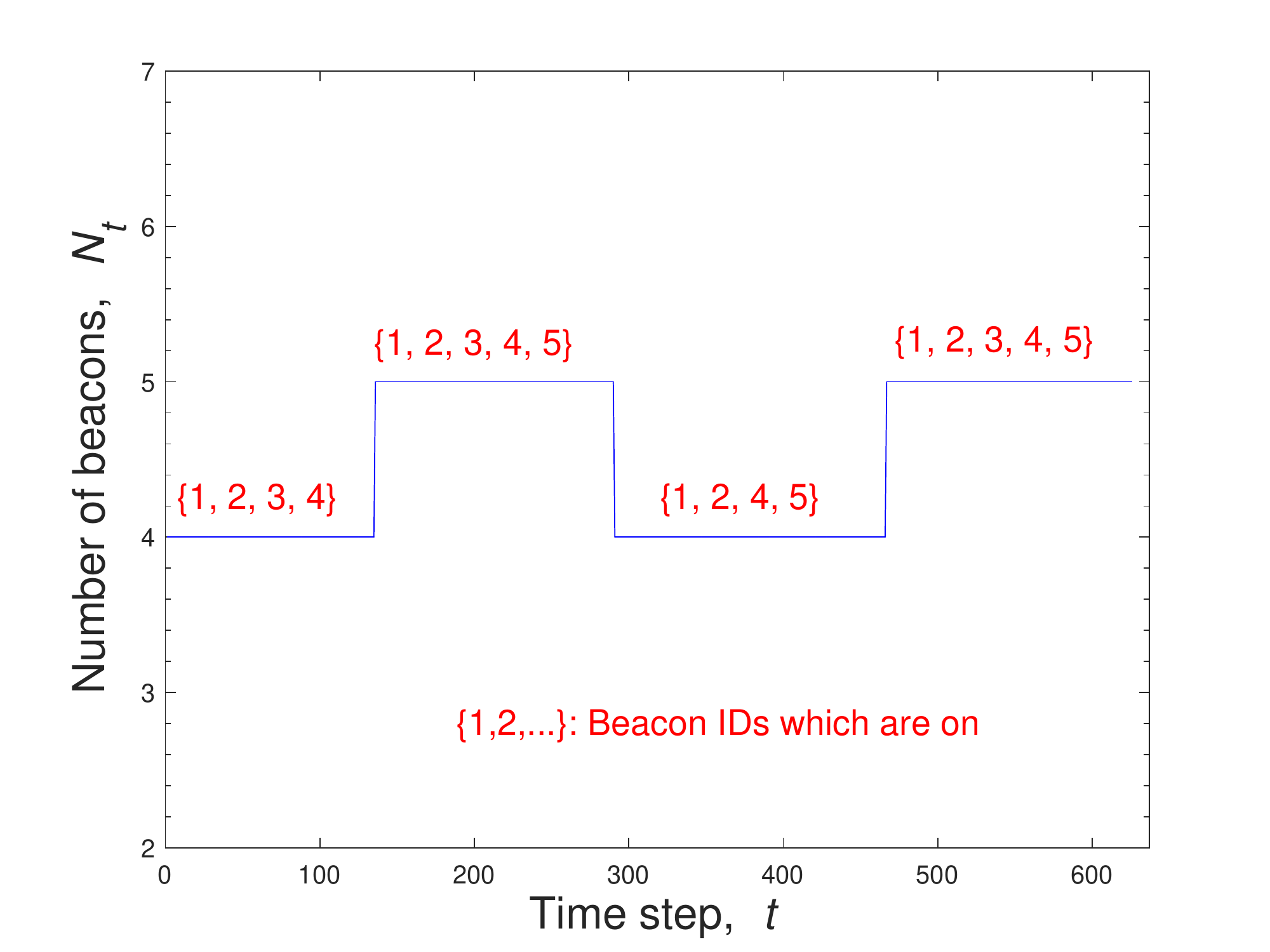}
\label{fig:ids}}}
\caption{Beacon(s) drops/joins/moves while SLAM is proceeding.}
\label{fig:dyn}
\vspace{-.15in}
\end{figure}

\begin{figure}[!htb]
\centering
\mbox{
\subfigure[\scriptsize{HectorSLAM at 0.4m/s}]{
\includegraphics[width=.28\linewidth]{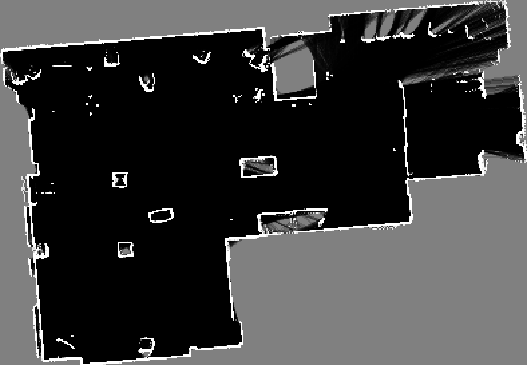}
\label{fig:h_slow}}
\hspace{-0.1in}
\subfigure[\scriptsize{HectorSLAM at 0.8m/s}]{
\includegraphics[width=.33\linewidth]{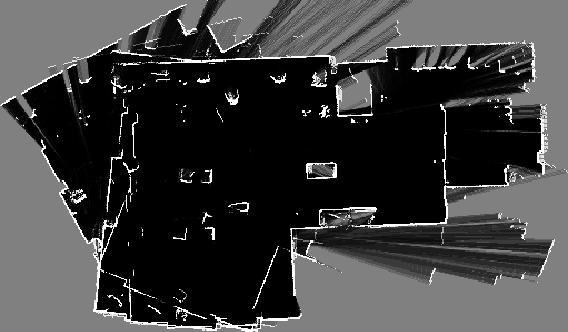}
\label{fig:h_fast}}
\hspace{-0.1in}
\subfigure[\scriptsize{UWB/LiDAR at 0.8m/s}]{
\includegraphics[width=.3\linewidth]{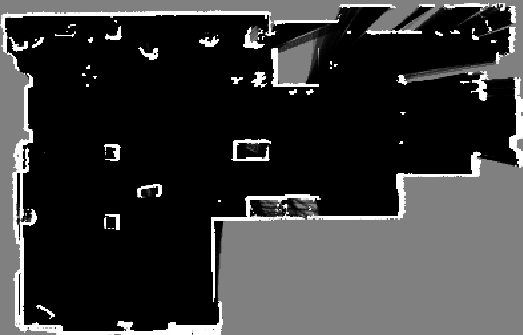}
\label{fig:p_fast}}
}
\caption{Comparison of our UWB/LiDAR-based SLAM with HectorSLAM \cite{kohlbrecher2011flexible} at different robot's speeds. To build the maps, the robot moves for one loop of the same trajectory as shown in \cref{fig:fusion1}.}
\label{fig:spd}
\vspace{-.2in}
\end{figure}

\cref{fig:dyn} demonstrates our system is capable of handling dynamical scenarios where the beacons may drop/join/move. We divide the SLAM process into three time slots: 
a) For $ t< 135$ (135 time steps not 135 seconds), there are four static beacons; 
b) We power on a new beacon (with ID. 5) at $t=135$  at the "start pos." shown in \cref{fig:slot1}, then this new beacon is moved along the trajectory of the robot until we place it to the "end pos." at $t=311$;
c) We power off one existing beacon (with ID. 3) at $t=311$ and relocate it to the "start pos." shown in \cref{fig:slot2}, then power it on at $t=466$ and move it along the trajectory of the robot until we place it to "end pos." at $t=626$. \cref{fig:ids} shows how the number beacons varies over time steps.
In \cref{fig:spd}, we compare UWB/LiDAR-based SLAM with HectorSLAM (LiDAR-only SLAM) when robot moves at different speeds. When the robot moves at about 0.4m/s, HectorSLAM is working properly and a high-quality map shown in \cref{fig:h_slow} is constructed. However, when the robot moves faster at about 0.8m/s,  HectorSLAM becomes error-prone where the errors mainly come from fast turnings as HectorSLAM cannot accommodate a sudden change of robot's behaviour especially the change of heading, and these errors would affect the subsequent scan matching steps thus being accumulated over time. This can be seen in \cref{fig:h_fast}.  Such errors can be dramatically reduced due to additional inputs from UWB sensors. In \cref{fig:p_fast} where robot moves as fast as the one in \cref{fig:h_fast}, we observe no drift-errors in UWB/LiDAR-based SLAM.

\subsection{SLAM in a corridor of length 22.7 meters}
\begin{figure}[!htb]
\vspace{-.1in}
\centering
\mbox{
\subfigure[HectorSLAM]{
\includegraphics[width=1\linewidth]{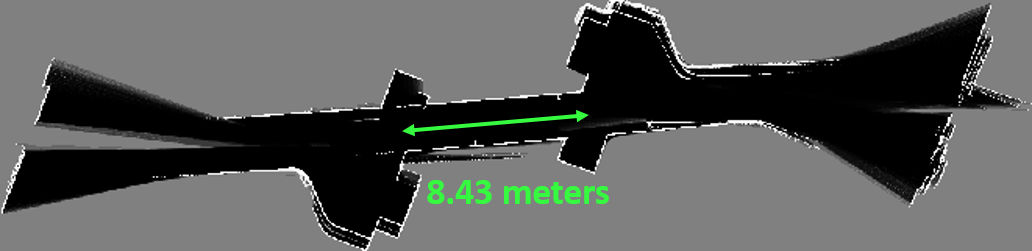}
\label{fig:f_hect}}}
\mbox{
\subfigure[UWB/LiDAR SLAM]{
\includegraphics[width=1\linewidth]{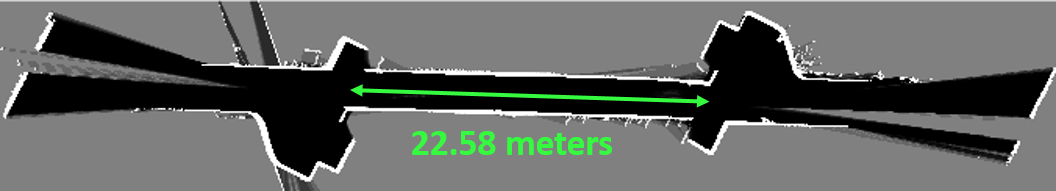}
\label{fig:f_prop}}
}
\caption{Comparison of our UWB/LiDAR-based SLAM with HectorSLAM \cite{kohlbrecher2011flexible} in a corridor of length 22.7m.}
\label{fig:feature}
\vspace{-.1in}
\end{figure}

HectorSLAM is known to work poorly in feature-less environments such as corridor and \cref{fig:f_hect} shows such an example where the length of corridor in its map is 8.43m whereas the actual length is 22.7m. Fusing LiDAR with UWB, we may regard UWB sensors as additional "features" that provide precise information about robot's location thus robustifying SLAM in such feature-less environments at the cost of slight degradation of map quality as shown in \cref{fig:f_prop}. 
\begin{Remark}
The quality of the map that UWB/LiDAR-based SLAM builds is compromised due to the tradeoff between UWB and LiDAR ranging accuracy which is reflected from \cref{eq:opt}. Moreover, to balance the UWB and LiDAR ranging accuracy, we add some Gaussian noise with zero mean and std. of 1cm/2cm to the laser ranger finder.
\end{Remark}

\section{Conclusion}
\label{section:c}
We have proposed a fusion scheme that utilizes both UWB and laser ranging measurements for simultaneously localizing the robot, the UWB beacons and constructing the LiDAR map. This is a 2D range-only SLAM and no control input is required. In our fusion scheme, a "coarse" map i.e. UWB map is built using UWB peer-to-peer ranging measurements. This "coarse" map can guide where the "fine" map i.e. LiDAR map should be built. Through a new scan matching procedure, the LiDAR map can be constructed while the UWB map is polished. 
Our proposed system suits infrastructure-less and ad-hoc applications where no prior information about the environment is known and the system needs to be deployed quickly.

\section*{Acknowledgments}
The work of Yang Song, Wee Peng Tay and Choi Look Law was partially supported by the ST Engineering - NTU Corporate Lab through the NRF corporate lab@university scheme Project Reference C-RP10B. The work of Mingyang Guan and Changyun Wen was partially supported by the ST Engineering - NTU Corporate Lab through the NRF corporate lab@university scheme Project Reference MHRP2.


\end{document}